\pdfoutput=1

\documentclass[11pt]{article}

\usepackage[]{ACL2023}

\usepackage{times}
\usepackage{latexsym}

\usepackage[T1]{fontenc}
\usepackage{fixltx2e}
\usepackage{subfigure}
\usepackage{tabularx}

\usepackage[utf8]{inputenc}

\usepackage{microtype}

\usepackage{inconsolata}

\title{Trusting Your Evidence:\\ Hallucinate Less with Context-aware Decoding}

\usepackage{fdsymbol}

\usepackage{xcolor}
\usepackage{times}
\usepackage{latexsym}
\usepackage{tabularx}
\usepackage{amsmath}
\usepackage{makecell}

\usepackage{longtable}

\usepackage[T1]{fontenc}
\PassOptionsToPackage{table}{xcolor}

\usepackage[utf8]{inputenc}

\usepackage{microtype}

\usepackage{inconsolata}
\newcommand{\ours}{CAD\xspace}

\usepackage{arydshln}

\usepackage{xcolor}

\usepackage{blindtext}
\usepackage{graphicx}
\usepackage{capt-of}
\usepackage{booktabs}
\usepackage{varwidth}
\usepackage{multirow}
\usepackage{xspace,mfirstuc,tabulary}

\newcolumntype{L}[1]{>{\raggedright\let\newline\\\arraybackslash\hspace{0pt}}m{#1}}
\newcolumntype{C}[1]{>{\centering\let\newline\\\arraybackslash\hspace{0pt}}m{#1}}

\newsavebox\tmpbox

\usepackage{bm}
\usepackage{color}
\usepackage{graphicx}
\usepackage[toc,page]{appendix}
\usepackage{makecell}
\usepackage{boldline}

\usepackage{algorithm}
\usepackage{colortbl}
\usepackage{tabstackengine}
\usepackage{tikz}
\usepackage{relsize}
\usepackage{microtype}
\usepackage{tablefootnote}
\usepackage{silence}
\usepackage{bbm}

\usepackage{tikz}
\usepackage{soul}

\usepackage{fdsymbol}

\definecolor{lightgray}{gray}{0.9}
\colorlet{soulgreen}{green!30}
\definecolor{red}{HTML}{FF0000}
\definecolor{blue}{HTML}{0000FF}
\definecolor{darkgreen}{HTML}{228B22}
\definecolor{dblue}{HTML}{007FFF}
\definecolor{figpurple}{HTML}{D5CEEF }
\definecolor{figblue}{HTML}{C9E3E5}
\definecolor{figred}{HTML}{E3B1B7}

\usepackage{booktabs,arydshln}

\makeatletter
\def\adl@drawiv#1#2#3{%
        \hskip.5\tabcolsep
        \xleaders#3{#2.5\@tempdimb #1{1}#2.5\@tempdimb}%
                #2\z@ plus1fil minus1fil\relax
        \hskip.5\tabcolsep}
\newcommand{\cdashlinelr}[1]{%
  \noalign{\vskip\aboverulesep
           \global\let\@dashdrawstore\adl@draw
           \global\let\adl@draw\adl@drawiv}
  \cdashline{#1}
  \noalign{\global\let\adl@draw\@dashdrawstore
           \vskip\belowrulesep}}
\makeatother

\usepackage[noend]{algpseudocode}
\algnewcommand{\parState}[1]{\State%
    \parbox[t]{\dimexpr\linewidth-\algmargin}{\strut\hangindent=\algorithmicindent \hangafter=1 #1\strut}}

\algrenewcommand\algorithmicindent{1.0em}%

\definecolor{magenta}{HTML}{F3DFF1}
\definecolor{red}{HTML}{FF0000}
\definecolor{hlgreen}{HTML}{D5E8D4}
\definecolor{figblue}{HTML}{DAE8FC}

\usepackage{algorithm}
\usepackage{colortbl}
\usepackage{tabstackengine}
\usepackage{tikz}
\usepackage{relsize}
\usepackage{microtype}
\definecolor{magenta}{HTML}{F3DFF1}
\definecolor{hlgreen}{HTML}{ccfcc4}
\definecolor{figblue}{HTML}{e7f2fe}

\usepackage{pifont}

\usepackage{paralist, tabularx}

\author{
 Weijia Shi $^{1}$ \thanks{\ \*  Equal contribution. Order randomly determined.} \qquad \qquad
    Xiaochuang Han $^{1}$ \footnotemark[1] 
        \\
        \textbf{Mike Lewis} $^{2}$ \quad 
        \textbf{Yulia Tsvetkov} $^{1}$ \quad
         \textbf{Luke Zettlemoyer} $^{1}$ \quad
        \textbf{Scott Yih $^{2}$} 
 \\
 \\ $^{1}$ University of Washington, Seattle, WA, $^{2}$ Meta AI \\
  {\tt \{swj0419, xhan77\}@cs.washington.edu}
}

\begin{document}
\maketitle

\setlength{\abovedisplayskip}{2pt}
\setlength{\belowdisplayskip}{2pt}
\begin{abstract} 
Language models (LMs) often struggle to pay enough attention to the input context, and generate texts that are unfaithful or contain hallucinations. To mitigate this issue, we present context-aware decoding (\ours), which follows a contrastive output distribution that amplifies the difference between the output probabilities when a model is used with and without context. 
Our experiments show that \ours, without additional training, significantly improves the faithfulness of different LM families, including OPT, GPT, LLaMA and FLAN-T5 for summarization tasks (e.g., 14.3\% gain for LLaMA in factuality metrics). Furthermore, \ours is particularly effective in overriding a model's prior knowledge when it contradicts the provided context, leading to substantial improvements in tasks where resolving the knowledge conflict is essential. 

\end{abstract}
\section{Introduction}

Language models (LMs) are remarkably effective in generating coherent and fluent continuations of a prompt or document prefix. During generation, they mostly rely on two sources of knowledge: (1) \textit{prior knowledge}, which is learned during pretraining and stored implicitly within the model parameters; (2) \textit{context knowledge}, which is passed as inputs in the prefix context \citep{Chan2022DataDP}. However, it remains an open question how a pretrained LM, particularly a vanilla LM without task-specific finetuning, balances these two knowledge sources during generation. 

\begin{figure}[]
    \centering
    \includegraphics[scale=0.295]{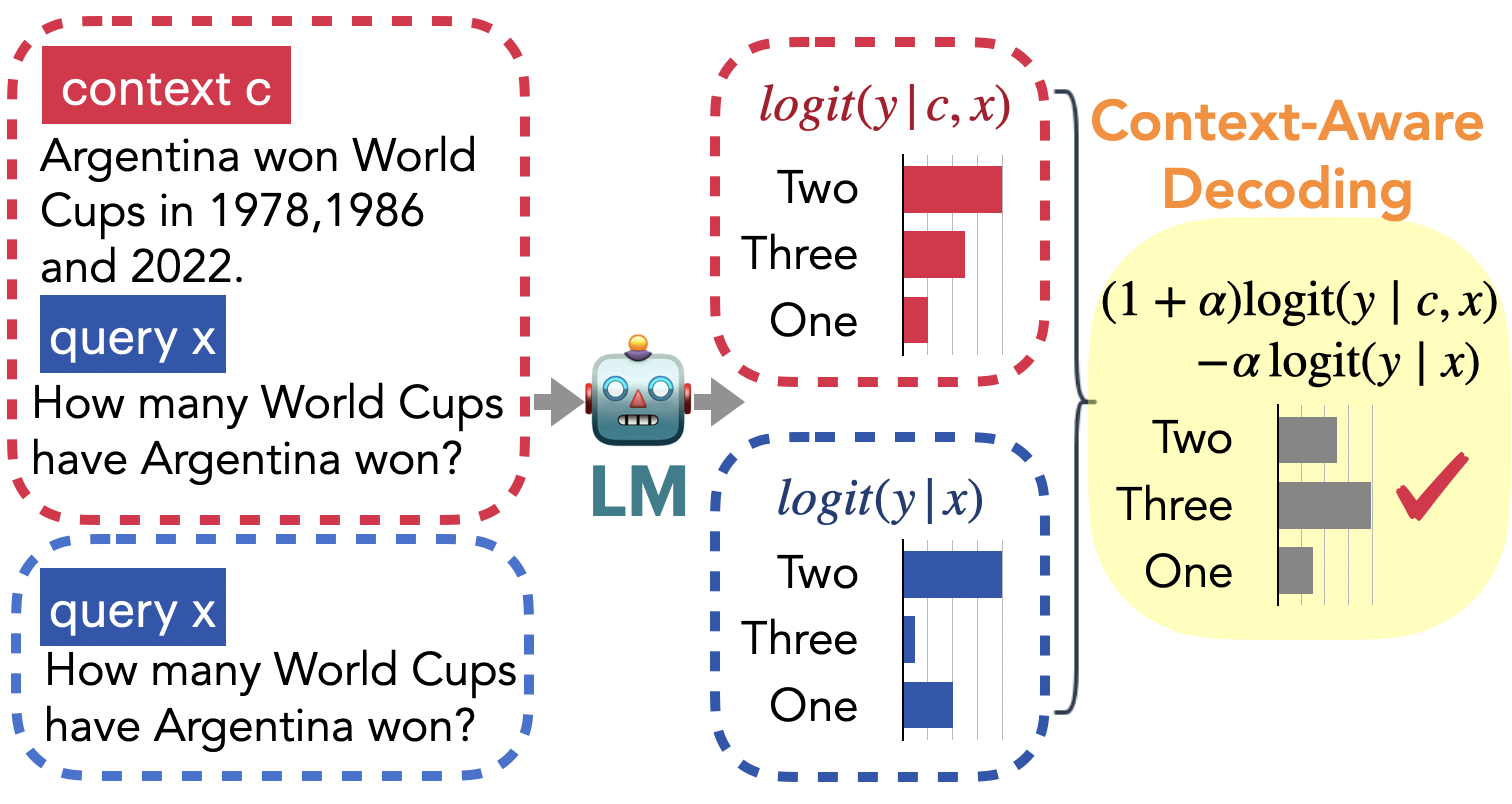}
    \caption{An illustration of context-aware decoding.} 
    \label{fig:intro}
\end{figure}

Previous research shows that LMs can fail to pay enough attention to new information introduced in the context knowledge. This can lead to hallucination in summarization~\cite{maynez-etal-2020-faithfulness, pagnoni-etal-2021-understanding}, where the generated summaries include facts not present in the input document. Insufficient attention to context is especially problematic when the context knowledge contradicts with the prior knowledge~\cite{longpre-etal-2021-entity, zhou2023contextfaithful}. 
For instance, when LLaMA~\cite{touvron2023llama} is presented with a latest document ``Argentina won the FIFA World Cups in 1978,1986 and 2022 ...'' in its context (\autoref{fig:intro}), it still predicts ``Two'' in response to the question ``How many World Cups have Argentina won?'', due in part to the outdated training data. 


In this work, we present a simple \textbf{c}ontext-\textbf{a}ware \textbf{d}ecoding (\ours{}) method to encourage the LM to attend to its context during generation.  As shown in \autoref{fig:intro}, \ours samples from a new output distribution, which amplifies the difference between output probabilities with and without the context document. This provides a new form of contrastive decoding~\citep{Li2022ContrastiveDO}, which effectively downweights the prior knowledge when more relevant contextual information is provided.  \ours can be used with off-the-shelf pretrained language models without any additional training. 

Experimental results from summarization tasks show that context-aware decoding significantly enhances the generation faithfulness of various vanilla LMs including OPT \cite{zhang2022opt}, GPT-Neo \cite{gpt-neo}, LLaMA \cite{touvron2023llama} and instruction-finetuned LMs such as FLAN~\cite{chung2022scaling}. For instance, when applied to LLaMA-30B in CNN-DM, CAD leads to substantial improvement in both \mbox{ROUGE-L}~(21\%) and summary factuality evaluation metrics~(14.3\%). 
More notably, \ours is especially beneficial for knowledge conflicting tasks, where the context contains information contradictory to the model's prior knowledge. \ours brings a 2.9x improvement to LLaMA-30B on a knowledge conflicts QA dataset~\cite{longpre-etal-2021-entity}. 
Furthermore, we observe that  this gain brought by \ours increases as the model size grows in knowledge conflicts tasks. 
These results demonstrate the potential of \ours in mitigating hallucinations in text generation and overriding prior knowledge with reliable and trusted information. 

\section{Method}

\subsection{Background}
Given a language model $\theta$, an input query $\boldsymbol{x}$, and a context $\boldsymbol{c}$ that contains some external knowledge \emph{unfamiliar} or \emph{in conflict} to the model's prior knowledge, we ask our model $\theta$ to generate a response $\boldsymbol{y}$ given the the query and context.
The response can be directly sampled (autoregressively) from the probability distribution conditioned on query $\boldsymbol{x}$ and context $\boldsymbol{c}$:  
\begin{align*}
    &y_t \sim p_\theta(y_t \mid \boldsymbol{c}, \boldsymbol{x}, \boldsymbol{y}_{<t}) \\
    &\propto \exp{\operatorname{logit}_\theta(y_t \mid \boldsymbol{c}, \boldsymbol{x}, \boldsymbol{y}_{<t})}
\end{align*}


However, in cases where the context $\boldsymbol{c}$ contains knowledge that is out-of-distribution with respect to $\theta$, we hypothesize that the model can struggle to effectively attend to $\boldsymbol{c}$ and overly rely on the prior knowledge encoded in $\theta$. For instance, as illustrated in Figure~\ref{fig:intro}, when the context $\boldsymbol{c}$ states ``Argentina won the FIFA World Cups in 1978, 1986 and 2022 ...'', it contradicts the LM's outdated prior knowledge that Argentina has won the World Cup twice. 
The language model may still incorrectly predict ``Two'' even when presented with the context $\boldsymbol{c}$ and the query $\boldsymbol{x}$.


\subsection{Context-aware Decoding}

To mitigate such issues, we factor out the prior knowledge from the model's original output distribution contrastively. Here, we model the prior knowledge as $p_{\theta}(y_t \mid \boldsymbol{x}, \boldsymbol{y}_{<t})$ and adjust the model's original output probability distribution using the pointwise mutual information (PMI) between the context $\boldsymbol{c}$ and the generation $y_t$, conditioned on $\boldsymbol{x}, \boldsymbol{y}_{<t}$. Formally, we have:
\begin{align*}
    &y_t \sim \Tilde{p}_\theta(y_t \mid \boldsymbol{c}, \boldsymbol{x}, \boldsymbol{y}_{<t}) \\
    &\propto p_\theta(y_t \mid \boldsymbol{c}, \boldsymbol{x}, \boldsymbol{y}_{<t}) \left( \frac{p_{\theta}(y_t \mid \boldsymbol{c}, \boldsymbol{x}, \boldsymbol{y}_{<t})}{p_{\theta}(y_t \mid \boldsymbol{x}, \boldsymbol{y}_{<t})} \right)^\alpha
\end{align*} 
where the output probability is a product-of-experts of the original output probability and PMI weighted by $\alpha$. Essentially, outputs that become much more likely when the context is included are preferred (Figure~\ref{fig:intro}). 

This expression is not a valid probability distribution and needs to be normalized across all possible values of $y_t$. By rearranging the terms, we obtain the final form:
\begin{align*}
    &y_t \sim \operatorname{softmax}[(1+\alpha) \operatorname{logit}_\theta(y_t \mid \boldsymbol{c}, \boldsymbol{x}, \boldsymbol{y}_{<t})\\
    &\qquad \qquad - \alpha \operatorname{logit}_\theta(y_t \mid \boldsymbol{x}, \boldsymbol{y}_{<t})]
\end{align*} \label{eq1}
Larger $\alpha$ means more weight on our adjustment ($\alpha=0$ reduces to regular decoding).\footnote{If we identify an external knowledge $\boldsymbol{c}$ conditionally independent to the generation, $p_{\theta}(y_t \mid \boldsymbol{c}, \boldsymbol{x}, \boldsymbol{y}_{<t}) = p_{\theta}(y_t \mid \boldsymbol{x}, \boldsymbol{y}_{<t})$, even a non-zero $\alpha$ would not have an impact to the original output distribution.}  We refer to this simple method as context-aware decoding. 
From the adjusted output distribution $\Tilde{p}$, we can apply various sampling strategies, such as nucleus sampling \cite{holtzmancurious}. 

Essentially, context-aware decoding is just a contrastive ensemble between the logits of $p_\theta(y_t \mid \boldsymbol{c}, \boldsymbol{x}, \boldsymbol{y}_{<t})$ and $p_\theta(y_t \mid \boldsymbol{x}, \boldsymbol{y}_{<t})$. 
A similar contrastive objective is universal in image generation, where classifier-free diffusion models \citep{Ho2022ClassifierFreeDG} predict diffusion noise with $(1 + \alpha) \boldsymbol{\epsilon}_\theta(\boldsymbol{x}, \boldsymbol{c}) - \alpha \boldsymbol{\epsilon}_\theta(\boldsymbol{x})$, with $\boldsymbol{c}$ being a control to the image. 
In text generation, \citet{Malkin2021CoherenceBW} propose coherence boosting with the same intuition, with a focus on contrasting the full input and a short premise-free input, promoting coherence w.r.t. the long context. 
Instead of using a single model $\theta$ in this work, different models can also be used in the distribution adjustments to demote unwanted model behaviors or distill expert model's capability \citep{Liu2021DExpertsDC,Li2022ContrastiveDO}.

\section{Experimental Setup}
We perform evaluation on tasks that require LMs to read and reason over contexts and produce outputs that are faithful to the contexts. Following prior work~\cite{zhang2023benchmarking, zhou2023contextfaithful}, we evaluate the models using prompting. 


\subsection{Datasets and Metrics}
\paragraph{Summarization} 
We conduct summarization experiments on two news datasets: CNN-DM~\cite{see2017get} and XSUM~\cite{narayan-etal-2018-dont}. We use ROUGE-L \cite{lin-2004-rouge} to evaluate summarization quality. To measure the factual consistency of summaries, we adopt BERT-Precision~\cite{pagnoni-etal-2021-understanding} as well as FactKB~\cite{feng2023factkb}, which has been demonstrated to achieve high correlations with human judgment on the two summarization datasets. 

\paragraph{Knowledge Conflicts}
We evaluate performance on two knowledge conflict datasets: MemoTrap \citep{memotrap} and NQ-Swap \cite{longpre-etal-2021-entity}. 
MemoTrap is created to investigate whether language models could fall into memorization traps. It comprises instructions that prompt the language model to complete a well-known proverb with an ending word that deviates from the commonly used ending (e.g., \emph{Write a quote that ends in the word ``early'': Better late than \rule{0.5cm}{0.15mm}}). 
NQ-Swap is based on a QA dataset, natural questions (NQ)~\cite{kwiatkowski2019natural}, where the objective is to answer questions based on a reliable gold document. To generate NQ-Swap, \citet{longpre-etal-2021-entity} first identify questions in NQ with named entity answers, find the supportive document for each question and then replace the gold answer entity in the document with a random entity. A faithful LM should generate the replaced entity as the answer when given the question and modified document. 
We also include the original NQ dataset with the question and original document for evaluation. We use Exact Match (EM) as the evaluation metric for NQ-Swap, NQ and MemoTrap. 

\medskip
In Table~\ref{tbl:setup}, we show illustrative examples of the contexts we aim to upweight for the model and the queries across different datasets. 
We hope LMs pay more attention to the source document in XSUM and NQ-Swap. On the other hand, we hope LMs focus more on the instruction in MemoTrap.

\begin{table}[t]
\small
\begin{tabularx}{0.48\textwidth}{lX}
\toprule
\multicolumn{2}{c}{\textbf{XSUM}} \\
\midrule
$\boldsymbol{c}$ &  \textcolor{red}{Article: Prison Link Cymru had 1,099 referrals in 2015-16 and said some ex-offenders were living rough for up to a year before finding suitable accommodation ...} \\
$\boldsymbol{x}$ & \textcolor{blue}{Summarize the article in one sentence. Summary:}  \\
\midrule
\multicolumn{2}{c}{\textbf{NQ-SWAP}} \\
\midrule
$\boldsymbol{c}$  & \textcolor{red}{Tesla CEO Elon Musk is now in charge of Twitter , CNBC has learned ...} \\
$\boldsymbol{x}$ & \textcolor{blue}{Who is Twitter CEO now?} \\
\midrule
\multicolumn{2}{c}{\textbf{MemoTrap}} \\
\midrule
$\boldsymbol{c}$ &  \textcolor{red}{Write a quote that ends in the word "early":} \\

$\boldsymbol{x}$ & \textcolor{blue}{Better late than} \\
\bottomrule
\end{tabularx}
\caption{An illustation of the inputs to \ours applied to each dataset. \ours upweights the context $\boldsymbol{c}$ (in red) by sampling each token from $ \operatorname{softmax}[ (1+\alpha) \operatorname{logit}_\theta(y_t \mid \boldsymbol{c}, \boldsymbol{x}, \boldsymbol{y}_{<t}) - \alpha \operatorname{logit}_\theta(y_t \mid \boldsymbol{x}, \boldsymbol{y}_{<t})]$. }
\label{tbl:setup}
\end{table}


\subsection{Models and Baselines}
We apply \ours to pretrained language models including OPT (13B and 30B)~\cite{zhang2022opt}, GPT-Neo (2.7B and 20B) \cite{gpt-neo}, LLaMA (13B and 30B) \cite{touvron2023llama} and instruction-finetuned language models such as FLAN-T5 (XL 3B and XXL 11B)~\cite{chung2022scaling}. 

\ours introduces a hyperparameter $\alpha$ to control the adjustment level. We set $\alpha=0.5$ for all models evaluated on the summarization datasets and $\alpha=1$ for all models evaluated on the knowledge conflict datasets. We observed that $\alpha=0.5$ generally yielded good results across all settings and all datasets, but a slightly higher $\alpha$ is more effective in the knowledge conflict setting, where the prior knowledge needs to be factored out more. We investigate the effect of $\alpha$ in Section \ref{sec:adjustment}. 

For the baselines, we use regular decoding following prior work~\cite{longpre-etal-2021-entity, kwiatkowski2019natural} to use greedy decoding for knowledge conflict tasks and top-$p$ sampling with $p$=0.9 for summarization tasks \citep{holtzmancurious}. 
For \ours, we use the same sampling strategies on top of the adjusted output probability distribution.

\begin{table*}[t]
\small
\centering
\begin{tabular}{llc|ccc|ccc}
\toprule
& & & \multicolumn{3}{c}{\textbf{CNN-DM}} & \multicolumn{3}{c}{\textbf{XSUM}} \\
\cmidrule(lr){4-6}
\cmidrule(lr){7-9}
\multicolumn{2}{c}{\textbf{Model}} & \textbf{Decoding} & ROUGE-L & factKB & BERT-P & ROUGE-L & factKB & BERT-P \\
\midrule
\multirow{4}{*}{ OPT } & \multirow{2}{*}{ 13B } & Regular & 22.0 & 77.8 & 86.5 & 16.4 & 47.2 & 85.2 \\
& & \ours & \textbf{27.4} & \textbf{84.1} & \textbf{90.8} & \textbf{18.2} & \textbf{64.9} & \textbf{87.5} \\
& \multirow{2}{*}{ 30B } & Regular & 22.2 & 81.7 & 87.0 & 17.4 & 38.2 & 86.1 \\
& & \ours & \textbf{28.4} & \textbf{87.0} & \textbf{90.2} & \textbf{19.5} & \textbf{45.6} & \textbf{89.3} \\
\midrule
\multirow{4}{*}{ GPT-Neo } & \multirow{2}{*}{ 3B } & Regular & 24.3 & 80.5 & 87.5 & 17.6 & 54.0 & 86.6 \\
& & \ours & \textbf{27.7} & \textbf{87.5} & \textbf{90.6} & \textbf{18.1} & \textbf{65.1} & \textbf{89.1} \\
& \multirow{2}{*}{ 20B } & Regular & 18.7 & 68.3 & 85.2 & 14.9 & 42.2 & 85.7 \\
& & \ours & \textbf{24.5} & \textbf{77.5} & \textbf{89.4} & \textbf{19.0} & \textbf{63.3} & \textbf{90.6} \\
\midrule
\multirow{4}{*}{ LLaMA } & \multirow{2}{*}{ 13B } & Regular & 27.1 & 80.2 & 89.5 & 19.0 & 53.5 & 87.8 \\
& & \ours & \textbf{32.6} & \textbf{90.8} & \textbf{93.0} & \textbf{21.1} & \textbf{73.4} & \textbf{91.7} \\
& \multirow{2}{*}{ 30B } & Regular & 25.8 & 76.8 & 88.5 & 18.7 & 47.7 & 87.1 \\
& & \ours & \textbf{31.8} & \textbf{87.8} & \textbf{92.2} & \textbf{22.0} & \textbf{66.4} & \textbf{90.3} \\
\midrule
\multirow{4}{*}{ FLAN } & \multirow{2}{*}{ 3B } & Regular & 25.5 & 90.2 & 91.6 & 18.8 & 31.9 & 88.2 \\
& & \ours & \textbf{26.1} & \textbf{93.9} & \textbf{92.1} & \textbf{19.5} & \textbf{35.9} & \textbf{88.8} \\
& \multirow{2}{*}{ 11B } & Regular & 25.4 & 90.4 & 91.4 & 19.4 & 29.8 & 88.3 \\
& & \ours & \textbf{27.1} & \textbf{93.1} & \textbf{92.2} & \textbf{20.0} & \textbf{35.0} & \textbf{88.8} \\
\bottomrule
\end{tabular}
\caption{\textbf{\ours consistently outperform the regular decoding method in terms of both summary quality metric (ROUGE-L) and summary factuality (factKB and BERT-P).} The best scores for each setting are boldfaced. FLAN 3B and 11B refer to FLAN-T5 XL and FLAN-T5 XXL respectively.
} \label{tbl:main}
\end{table*}

\begin{table}[t]
\small
\begin{tabular}{p{0.85cm}p{0.45cm}c|ccc}
\toprule
\multicolumn{2}{c}{\textbf{Model}} & \textbf{Decoding} & \textbf{Memo.} & \textbf{NQ} & \textbf{NQ-SWAP} \\
\midrule
\multirow{4}{*}{ OPT } & \multirow{2}{*}{ 13B } & Reg. & 32.5 & 29.2 & 18.8 \\
& & \ours & 44.5 & 32.2 & 36.9 \\
& \multirow{2}{*}{ 30B } & Reg. & 28.4 & 29.4 & 14.7 \\
& & \ours & 41.0 & 35.5 & 29.0 \\
\midrule
\multirow{4}{*}{ GPT. } & \multirow{2}{*}{ 3B } & Reg. & 22.5 & 31.9 & 19.1 \\
& & \ours & 47.3 & 39.9 & 41.2 \\
& \multirow{2}{*}{ 20B } & Reg. & 37.1 & 22.8 & 16.1 \\
& & \ours & 57.3 & 32.1 & 36.8 \\
\midrule
\multirow{4}{*}{ LLAMA } & \multirow{2}{*}{ 13B } & Reg. & 23.8 & 22.3 & 11.7 \\
& & \ours & 57.1 & 33.6 & 36.7 \\
& \multirow{2}{*}{ 30B } & Reg. & 25.8 & 23.8 & 9.6 \\
& & \ours & 50.6 & 34.0 & 37.7 \\
\midrule
\multirow{4}{*}{ FLAN } & \multirow{2}{*}{ 3B } & Reg. & 69.2 & 81.8 & 71.4 \\
& & \ours & 72.2 & 80.3 & 73.3 \\
& \multirow{2}{*}{ 11B } & Reg. & 82.0 & 85.5 & 73.0 \\
& & \ours & 88.7 & 82.5 & 77.1 \\
\bottomrule
\end{tabular}
\caption{\ours outperforms the regular decoding method (Reg.) in all settings except for FLAN-T5 on NQ. Note that FLAN-T5 is trained on NQ dataset during instruction-finetuning. 
} \label{tbl:nq}
\end{table}

\section{Results}
\subsection{Main Results}
\paragraph{Summarization}
Table~\ref{tbl:main} reports the results on CNN-DM and XSUM. We observe that \ours outperforms the standard decoding algorithm by a large margin in all eight models across both datasets. Specifically, when applied to LLAMA-30B in CNN-DM, \ours leads to 21\% increase in ROUGE-L, 14.3\% increase in factKB and 7.8\% increase in BERT-P. 
This result demonstrates that \ours could effectively improve the quality and factuality of the generated summaries from a diverse set of language models. 

\paragraph{Knowledge Conflicts}
Our results for the knowledge conflict datasets, NQ-SWAP and MemoTrap, as well as the original NQ are detailed in Table~\ref{tbl:nq}. \ours is significantly better than the regular decoding in all settings, with the exception of a minor decrease observed for FLAN-T5 on the non-conflict NQ dataset.\footnote{This slight decline can be attributed to the fact that this particular NQ dataset is included in the instruction-finetuning sets used by FLAN-T5, and hence, the model has been previously trained on it.} Despite this, \ours achieves substantially better performance on the knowledge conflict datasets, e.g., \ours improve GPT-Neo 20B by 54.4\% on Memotrap and by 128\% on NQ-SWAP. 
This substantial improvement suggests that context-aware decoding is particularly beneficial for LMs to adhere to the given context, in scenarios where the model's prior knowledge contradicts with the context knowledge. 


\subsection{Analysis}
\paragraph{Qualitative analyais}
\begin{table}[h!]
\small
\begin{tabularx}{0.5\textwidth}{lX}
\toprule
\multicolumn{2}{c}{\textbf{XSUM}} \\
\midrule
Article & He passed away peacefully in hospital on Tuesday after a short illness. Born in Tourmakeady, County Mayo, he worked as a teacher before securing a part in the premiere of the Brian Friel play Translations in 1980. Lally became a household name in Ireland for his role as Miley Byrne in the RTE soap opera Glenroe and later starred in the BBC series Ballykissangel. He also appeared in the Hollywood movie Alexander and provided the voice for the Oscar-nominated, animated Irish film, The Secret of Kells. As a fluent Irish speaker and advocate of the language, Lally had roles in several Irish language films ... \\
Regular & \colorbox{yellow}{Westminister actor Pat} Lally died in hospital on Tuesday night \colorbox{yellow}{aged 82}  \\
\ours & Actor Lally, best known for Glenroe and Ballykissangel, has died in hospital on Tuesday \\
\midrule
\multicolumn{2}{c}{\textbf{MemoTrap}} \\
\midrule
Input &  Write a quote that ends in the word ``early''. Better late than \\
Regular & \colorbox{yellow}{never} \\
\ours & early \\
\bottomrule
\end{tabularx}
\caption{Qualitative examples of contrast-aware decoding. The nonfactual or inconsistent texts are highlighted in yellow. }. \label{tbl:qualitative}
\end{table}
We provide qualitative examples for XSUM and Memotrap in Table~\ref{tbl:qualitative}. In XSUM, the regular decoding generates texts that is not mentioned in the article, whereas \ours produces output exclusively based on the information in the input article. 
For MemoTrap, the standard decoding disregards the instruction and generates the memorized ending, while \ours adheres to the instruction within the given context and produces the desired output. 
\begin{figure*}[ht]
    \centering
    \includegraphics[scale=0.5]{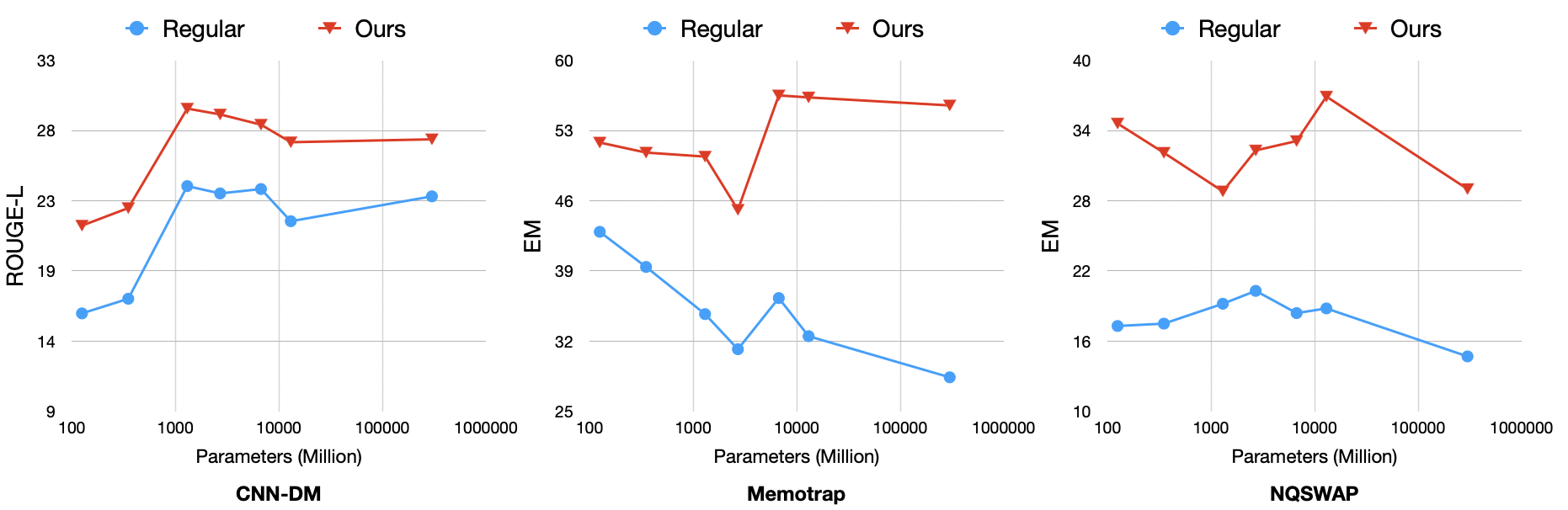}
    
    \caption{OPT models of varying sizes consistently benefit from \ours. The x-axis indicates the size of language models and the y-axis is the performance. } \label{fig:size}
\end{figure*}

\paragraph{\ours brings consistent improvement to LMs with different sizes.}
In Tables~\ref{tbl:main} and~\ref{tbl:nq}, we show that \ours could be used to enhance a diverse set of LM families, including OPT, GPT-Neo, LLaMA, and FLAN-T5. Here we further investigate whether \ours is effective in improving language models of different sizes. Specifically, we focus on OPT models across a range of sizes: 125M, 350M, 1.3B, 2.7B, 6.7B, 13B, 30B. As depicted in Figure~\ref{fig:size}, we observe that the performance gain brought by \ours stays consistent with 
different model sizes in CNN-DM. In Memotrap and NQSWAP, this gain increases as the model size grows, indicating that larger LMs can have a greater tendency to rely on their prior knowledge instead of reading the contexts, thereby benefiting more from \ours.

\paragraph{Effect of adjustment level $\alpha$} 
\label{sec:adjustment}

\begin{figure*}[ht]
    \centering
    \includegraphics[scale=0.4]{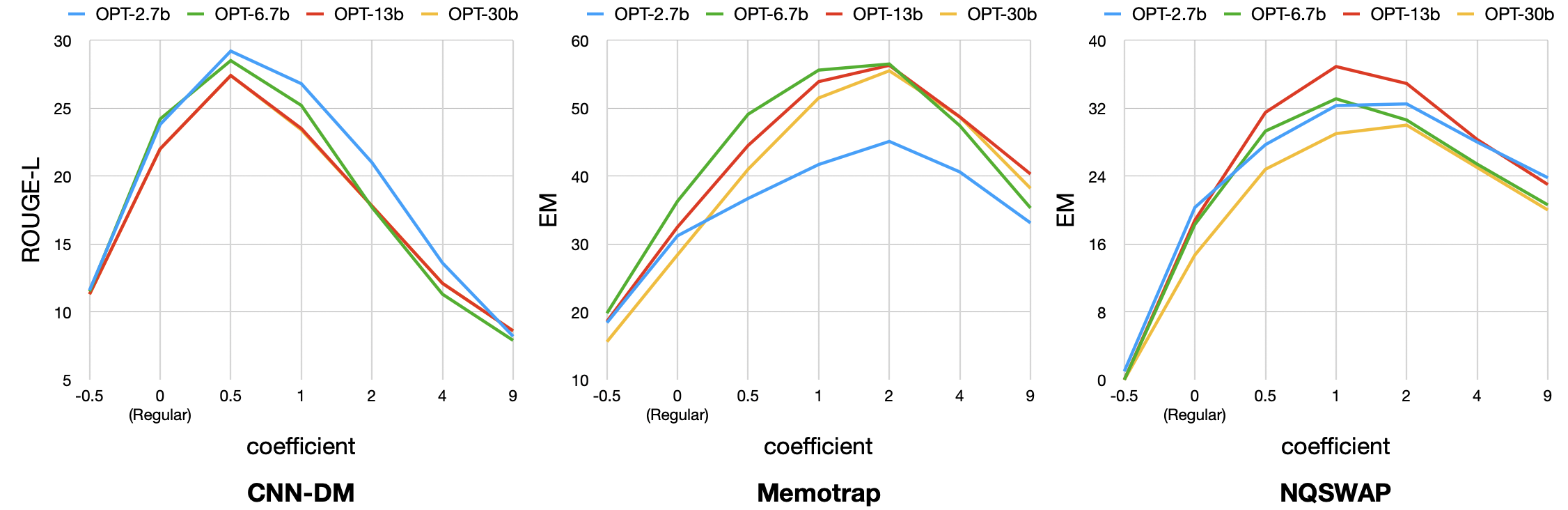}
    \caption{Effect of the adjustment level $\alpha$. The y-axis is the performance and the x-axis is $\alpha$.} 
    \label{fig:lambda}
\end{figure*}

Context-aware decoding introduces a hyperparameter $\alpha$, which serves to control the adjustment level of \ours (a small $\alpha$ makes the distribution closer to the original next token distribution). 
We conduct experiments with various values of $\alpha$ and present the results in \autoref{fig:lambda}. Across all three datasets, we find $\lambda=0.5$ consistently provide robust improvements over regular decoding. Further increasing the value of $\alpha$ yields additional improvement in tasks involving knowledge conflicts.

\section{Related Work}
\paragraph{Summarization Factuality}
Summarization models have shown a tendency to generate hallucinated texts~\cite{maynez-etal-2020-faithfulness, pagnoni-etal-2021-understanding}. This has led to growing efforts to improve the factual consistency, including applying attentions to fact triples extracted from source documents \cite{cao2018faithful, zhu-etal-2021-enhancing}, optimizing summarization models towards a factual consistency metrics 
\cite{nan-etal-2021-improving, cao2021cliff}, learning a post-editing error corrector~\cite{dong-etal-2020-multi} and removing noisy training samples \cite{kang-hashimoto-2020-improved, goyal-durrett-2021-annotating}. However, all these methods require additional fine-tuning and are not directly suitable for zero-shot and few-shot prompting scenarios. 

\paragraph{Knowledge Conflicts}

When presented with an updated document with conflicting knowledge, we expect language models to generate responses based on the provided contexts rather than relying solely on outdated parametric knowledge. 
%
%
This setting is especially valuable to retrieval-augmented language models~\cite{Khandelwal2020Generalization, shi2023replug, min2022nonparametric, yasunaga2022retrievalaugmented}, where documents retrieved from external databases are used as additional input to provide LMs additional knowledge. 
However, simply adding documents does not always change the model predictions, as current LMs often overlook the contexts and rely heavily on their prior parametric knowledge~\cite{longpre-etal-2021-entity, chen-etal-2022-rich}.
Existing approaches for improving model's faithfulness to the context, such as the prompting-based method~\cite{zhou2023contextfaithful}, are limited in that they could only apply to large-scale instruction-finetuned LMs like OpenAI's text-davinci-003. In contrast, our work investigates a decoding strategy to tackle this problem, which is applicable to any LM.



\paragraph{Contrastive Decoding Methods}
Contrastive decoding methods have been extensively explored for text generation. 
Coherence boosting \citep{Malkin2021CoherenceBW} demotes a short context from a full context, focusing on the longer-range context for coherence and overall better generation quality. MMI-based decoding \citep{Li2015ADO} uses a contrastive formulation to improve output diversity in dialog generation. 
In this work, we adopt a same intuition and focus on analyzing the knowledge conflict scenarios where the faithfulness to the context is particularly important but difficult for the regular decoding methods. 
DExperts \citep{Liu2021DExpertsDC} demotes the output distribution of an \emph{anti}-expert (e.g., exposed to toxic language) to help lead the generations free from the unwanted attributes. Contrastive decoding \citep{Li2022ContrastiveDO} demotes an \emph{amateur} model (e.g., models with a very small number of parameters) to help distill the expert knowledge learned in the larger, more competitive models. 
In general, contrastive decoding has shown to be a general way to control model outputs, which we reinforce by considering the new case of factual consistency with the textual context. 

\section{Conclusion}
Off-the-shelf language models may suffer from an insufficient attention to the supplied context compared to its learned prior knowledge, leading to an unfaithful generation to the input context. We present context-aware decoding, a simple inference-time method that downweights an output probability associated with the model's prior knowledge to promote models' attention to the contextual information. We experiment on two families of tasks that require a strong attention to the context, summarization and knowledge conflicts tasks. We show that \ours provides more reliable and factual outputs across different language models of various sizes. 


\bibliography{anthology,custom}
\bibliographystyle{acl_natbib}

\appendix
\end{document}